\documentclass[runningheads]{llncs}
\usepackage{caption}
\usepackage{arydshln}
\usepackage{tabularray}
\usepackage{siunitx}        
\usepackage{etoolbox} 
\usepackage{graphicx}
\usepackage{amsmath}
\usepackage{amssymb}
\usepackage{booktabs}
\usepackage{makecell}
\usepackage{bbm}
\usepackage{bm}
\usepackage{multirow}
\usepackage{verbatim}
\usepackage[ruled]{algorithm2e}
\usepackage{colortbl}
\usepackage[dvipsnames]{xcolor}
\usepackage[colorlinks,linkcolor=blue,citecolor=blue]{hyperref}
\usepackage{booktabs,subcaption,amsfonts,dcolumn}
\providecommand{\zkreffig}[1]{Figure~\ref{#1}} 
\providecommand{\zkreftb}[1]{Table~\ref{#1}}
\begin{document}
\title{ModelMix: A New Model-Mixup Strategy to Minimize Vicinal Risk across Tasks for Few-scribble based Cardiac Segmentation}
\titlerunning{ModelMix}
%
 \author{Ke Zhang \and  Vishal M. Patel}
\authorrunning{K Zhang \& VM Patel}
%
\institute{Department of Electrical and Computer Engineering, The Johns Hopkins University
\email{kzhang99@jhu.edu}}
\maketitle              

\begin{abstract}
Pixel-level dense labeling is both resource-intensive and time-consuming, whereas weak labels such as scribble present a more feasible alternative to full annotations.
However, training segmentation networks with weak supervision from scribbles remains challenging. 
Inspired by the fact that different segmentation tasks can be correlated with each other, we introduce a new approach to few-scribble supervised segmentation based on model parameter interpolation, termed as \emph{ModelMix}.
Leveraging the prior knowledge that linearly interpolating convolution kernels and bias terms should result in linear interpolations of the corresponding feature vectors, ModelMix constructs virtual models using convex combinations of convolutional parameters from separate encoders.
We then regularize the model set to minimize vicinal risk across tasks in both unsupervised and scribble-supervised way.
Validated on three open datasets, \emph{i.e.}, ACDC, MSCMRseg, and MyoPS, our few-scribble guided ModelMix significantly surpasses the performance of the state-of-the-art scribble supervised methods.
\end{abstract}
\section{Introduction}
Creating large-scale fully annotated medical image datasets is both time-consuming and burdensome. To address this bottleneck, researchers have investigated weak annotations~\cite{tajbakhsh2020embracing}, such as image-level labels, sparse labels, and noisy labels. 
Among these alternatives, scribble is a particularly attractive choice due to its advantages in annotating complex structures~\cite{Can2018LearningTS}.
We propose to explore few-shot scribble supervised segmentation, which further reduces the annotation effort by exploiting several scribble annotated images and a large amount of unlabeled images.                                                              

Few-shot scribble supervised segmentation is particularly challenging due to the scarcity of annotations.
Existing methods mainly exploit the labeled pixels~\cite{bai2018recurrent,Can2018LearningTS,ji2019scribble,l2016inscribblesup,luo2022scribble} and regularize the model training with priors~\cite{9389796,zhang2022shapepu,zhang2023zscribbleseg}.
These methods are susceptible to the supervision amount and might easily over-fit to annotated samples.

Several studies have been conducted to explore augmentation techniques based on mixup, including Mixup~\cite{zhang2018mixup}, Cutout~\cite{devries2017cutout}, Cutmix~\cite{yun2019cutmix}, PuzzleMix~\cite{kimICML20} and Comixup~\cite{kim2021comixup}. These works interpolate images to generate unseen samples. 
However, it is difficult to directly interpolate labels for different datasets due to disparity in label categories. Meanwhile, considering the domain gap, directly blending images from distinct segmentation tasks may lead to unrealistic results.

To address the above challenges, we propose to mix model parameters of complementary tasks. Our method is based on two assumptions:  (1) we assume that different segmentation tasks are intrinsically related, such as myocardial pathology and cardiac structure segmentation.
(2) we suppose the individual models for each task are distributed in the vicinity of the general model. 
One direct solution is to train a general model for correlated tasks.
However, the model is faced with the trade-off between learning domain-invariant knowledge and domain-specific information, especially when the model size is not large enough.
To tackle this, we propose to train individual models for each task, while regularize them with vicinal principle. Specifically, we construct virtual models from the individual models, and then apply vicinal regularization to the virtual models.

We propose to mix model parameters learned from correlated tasks, termed as ModelMix, for few-shot scribble-guided cardiac segmentation.
Firstly, we train individual models for each task and employ techniques such as data augmentation and dropout to enhance the network's robustness.
Secondly, we conduct linear interpolation between randomly selected convolution parameters from the encoders of separate models to construct virtual networks.
Thirdly, we apply vicinal constraints between the hybrid and non-hybrid versions of the model to encourage the mixed networks to generalize across all individual tasks.

The contributions of this work are two-fold. (1) We propose a holistic ModelMix strategy to learn from complementary segmentation tasks. Specifically, we construct mixed models and regularize them to have consistent performance with each individual model. (2)  Evaluated on three public datasets of ACDC, MSCMRseg, and MyPS, our ModelMix demonstrated its advantage over existing scribble supervised segmentation approaches.

\begin{figure}[t]
    \centering
    \includegraphics[width=0.9\textwidth]{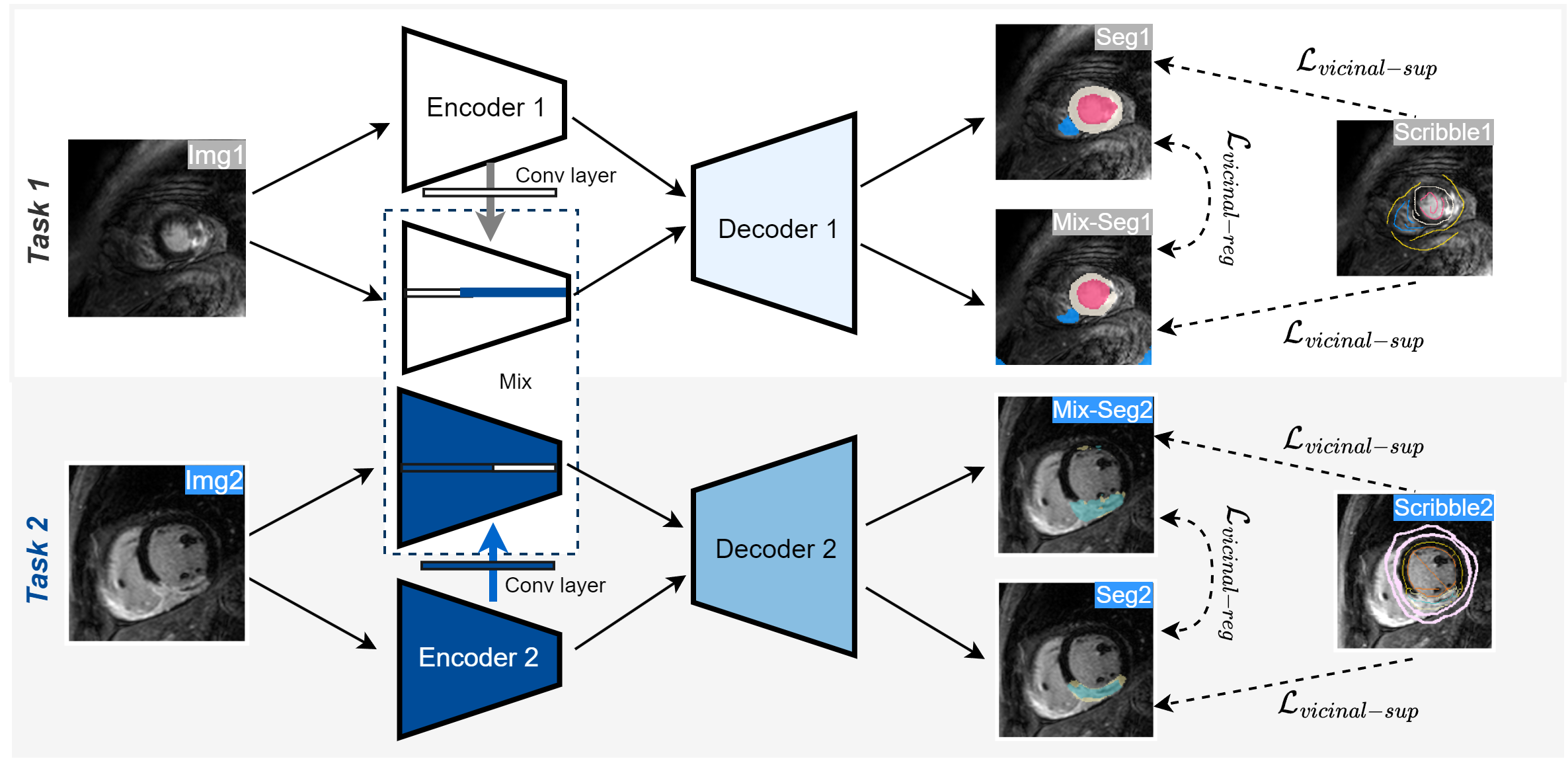}
    \caption{Overview of the proposed ModelMix framework for cardiac segmentation from scribble supervision.}
    \label{fig:pipeline}
\end{figure}

\section{Method}
As shown in \zkreffig{fig:pipeline}, our ModelMix linearly interpolates the parameters of encoder convolutional layer for different tasks, and requires the segmentation of mixed model and individual model to be consistent. 
The proposed ModelMix is composed of three components: (1) The individual models are trained end-to-end for different tasks.
(2) For the encoder trained for a separate task, we randomly select the convolutional layers and mix the parameters with linear interpolation while keeping the parameters of other layers unchanged.
(3) Both supervised and unsupervised vicinal regularization are applied to require consistent segmentation results between hybrid and non-hybrid models.

\subsection{Train for separate tasks}
For $n$ segmentation tasks $\{t_i\}_{i=1}^n$, we adopt the same encoder architecture and separate segmentation decoder, denoted as $\{e_1, e_2, \cdots, e_n\}$ and $\{h_1,h_2,\cdots, h_n\}$, respectively. 
The encoders have the same architecture but do not share model weights.
To improve the robustness of neural network, we introduce both image- and model-level perturbation to the training process.

For perturbations at the image level, we conduct enhancement operations both within individual images and between different images.
For each individual image, we randomly cutout a square area of pixels. 
Let $x$ denote the input image and $\mathbbm{1}_c$ be the binary cutout mask, the obtained results is represented as $x^c = \mathbbm{1}_c\odot x$.
For perturbation across images, we perform linear interpolation between images from the same dataset.
Let $\alpha$ be the mix ratio sampled from beta distribution, we define the perturbed image as $x' = \alpha x^c_1 +(1-\alpha) x^c_2$. Inspired by~\cite{zhang2022cyclemix}, we require the mix-equivalence of image segmentation with invariant loss $\mathcal{L}_{\text{inv}}$:
\begin{equation}
\mathcal{L}_{\text{inv}} = \mathcal{L}_{\text{cos}}\left(f(x'), \alpha f(c_1\cdot x_1)+(1-\alpha)f(c_2\cdot x_2)\right).
\end{equation}
For model-level perturbation, we simply apply dropout to the encoder of each separate model.
The design serves to mitigate over-fitting and improve the robustness of neural networks.

\subsection{Minimize vicinal risk across tasks}
For segmentation tasks, our objective is to minimize the mean of the loss function $l$ across the distribution of models $P$, referred to as the expected risk. Empirically, we approximate the model distribution using the model set $\{f_i\}_{i=1}^n$:
\begin{equation}
R_\delta(f) = \int l(f_i(x),y)dP_\delta(f) = \frac{1}{n}\sum_{i=1}^n l(f_i(x),y),
\end{equation}
where $\delta(f_i)$ is the Dirac mass centered at $f_i$. According to the vicinal risk minimization principle~\cite{chapelle2000vicinal}, the distribution of function $f$ is extended to:
\begin{equation}
P_v(\tilde{f}) = \frac{1}{n}\sum_{i,j} v(\tilde{f}_{ij}|f_i,f_j ).
\end{equation}
We propose a generic vicinal distribution by \emph{ModelMix}.
For complementary tasks pair set $\Omega$, we sample the task pair $(t_i,t_j)\in \Omega$.
We assume different tasks share the general encoder and separate segmentation decoder, \emph{i.e.}, $f_i(x) = h_i(\bar{e}(x))$
Then, we construct the general vicinal distribution as follows:
\begin{equation}
\mu(\tilde{f}|f_i,f_j) = \frac{1}{n}\sum_{i,j} E[\delta(\tilde{f}_{ij} = h(\tilde{e}) | f_i,f_j)],
\end{equation}
where $\tilde{e}$ denotes the constructed virtual encoder. We assume $e_i$ is distributed in the vicinity of the general backbone $\bar{e}$ for each separate task $t_i$. Then, we simply extend the vicinal distribution of $\tilde{e}$ by incorporating the prior knowledge that linear interpolation of the convolution parameters results in linear interpolation of the relevant features.
Taking the convolution operation as $g(\bm{x},\bm{k},b)$ with kernel $\bm{k}$ and bias $b$, the constructed convolution $\tilde{g}(\bm{x},\bm{k},b)$ is defined as:
\begin{gather}
\begin{aligned}
\tilde{g}_{ij}(\bm{x},\bm{k},b) &= \lambda g_i(\bm{x},\bm{k},b) +(1-\lambda) g_j(\bm{x},\bm{k},b)
\\
&= \bm{x}*\left[\lambda\bm{k}_i + (1-\lambda) \bm{k}_j\right] +\lambda \left[b_i+(1-\lambda)b_j\right],
\end{aligned}
\end{gather}
where $\lambda$ is the mixed ratio sampled from beta distribution.
For each encoder, we randomly select a convolutional layer from the encoder to perform interpolation between tasks and keep the parameters of other layers unchanged.

 \subsection{Vicinal regularization}
We apply vicinal regularization in both unsupervised and scribble-supervised manner to enable the constructed virtual model have consistent performance with individual models on each single task.
Firstly,  we regularize the output of constructed model $\tilde{f}_{ij}$ to be consistent with $f_i$:
\begin{gather}
 \mathcal{L}_{\text{vicinal-reg}}= \sum_{i,j}\mathcal{L}_{\text{cos}}(\tilde{f}_{ij}(x), f_i(x)) = -\frac{\tilde{f}_{ij}(x) \cdot f_i(x)}{ \|\tilde{f}_{ij}(x)\|_2 \cdot \|f_i(x)\|_2}, 
\end{gather}
which minimizes the cosine similarity between $\tilde{f}_{ij}(x)$ and $f_i(x)$.
Secondly, we leverage the supervision of scribble annotations, and calculate the supervised loss for $\tilde{f}_{ij}$, $f_i$, respectively:
\begin{gather}
    \mathcal{L}_{\text{vicinal-sup}}  = \sum_{i,j} \left[\mathcal{L}_{\text{sup}}(y,\tilde{f}_{ij}(x)) +  \mathcal{L}_{\text{sup}}(y,f_{i}(x)) \right]\\
        \mathcal{L}_{\text{sup}}(y,f(x)) = -\left[y\log(f(x)) +\frac{2yf(x)}{y+f(x)}\right],
\end{gather}
where $\mathcal{L}_{\text{sup}}$ is the combination of cross entropy and Dice loss calculated for annotated pixels. 
Then, our overall training objective is formulated as:
\begin{equation}
\mathcal{L} = \mathcal{L}_{\text{inv}} + \mathcal{L}_{\text{vicinal-reg}} + \mathcal{L}_{\text{vicinal-sup}}.
\end{equation}

\section{Experiment}
\subsection{Experiment Setup}
\textbf{Datasets:} 
 \textbf{ACDC}\footnote{\url{https://www.creatis.insa-lyon.fr/Challenge/acdc/databasesTraining.html}} includes cardiac MR images obtained from 100 patients, with the annotations of the right ventricle (RV), left ventricle (LV), and myocardium (MYO). Following~\cite{zhang2022shapepu}, we partitioned the 100 subjects into three groups: a training set comprising 70 subjects, a validation set with 15 subjects, and a test set of 15 subjects. For weak-supervision studies, we utilized expert manually annotated scribble annotations released by \cite{9389796}.
\textbf{MSCMRseg}\footnote{\url{http://www.sdspeople.fudan.edu.cn/zhuangxiahai/0/mscmrseg19/data.html}} consists of late gadolinium enhancement (LGE) cardiac MR images from 45 patients diagnosed with cardiomyopathy. We adopt scribble annotations from~\cite{zhang2022cyclemix}. Following ~\cite{gao2023bayeseg}, we divide the dataset randomly into 25 training images, 5 validation images, and 15 test images for evaluation.
\textbf{MyoPS}\footnote{\url{https://zmiclab.github.io/zxh/0/myops20/}} comprises 45 paired multi-sequence CMR images encompassing Balanced Steady-State Free Precession (BSSFP), Late Gadolinium Enhancement (LGE), and T2-weighted CMR sequences. Compared to the structural segmentation task, MyoPS is more challenging due to the diverse representation of pathology across different patients, posing difficulties in portraying scar and edema characteristics. We leverage the scribbles released by~\cite{zhang2023zscribbleseg} and divided the dataset into 20 pairs for training, 5 for validation, and 20 for testing following~\cite{li2023myops}. \emph{For all datasets, we train the models using five randomly selected scribble-annotated volumes along with other unlabeled images in the training set.}
\\
\\
\noindent\textbf{Implementation:} 
We utilized the UNet~\cite{ronneberger2015u} architecture as the foundational structure for our segmentation network. To introduce perturbations and enhance training, a dropout layer (with a ratio of 0.5) was incorporated before each convolutional block.  We apply random rotations, flips, and crop to augment the training dataset, with resulting images resized to 256 × 256 for network input. The intensity of each slice is scaled to a range of 0-1. All models are trained with batch size of 16 and a learning rate of 1e$^{-3}$. We use evaluation metrics of Dice scores and Haussdorff Distance (HD) to assess segmentation accuracy. Implementation was conducted over 1000 epochs on eight NVIDIA 3090Ti 24GB GPUs.

\begin{table}[t]
	\caption{Component ablations: 5-scribble based segmentation results in Dice scores of the ablation study. \textbf{Bold} denotes the best result, \underline{underline} indicates the  best but one. }
	\begin{center}
		\resizebox{\textwidth}{!}{
		\begin{tabular}{ccccccccccccccccc}
         \hline
	 \hline
	\multirow{2}{*}{Methods}&\multirow{2}{*}{$\mathcal{L}_{\text{pce}}$}&\multirow{2}{*}{$\mathcal{L}_{\text{inv}}$}& \multicolumn{2}{c}{$\mathcal{L}_{\text{vicinal}}$} &\multirowcell{2}{Share\\ Enc} & \multicolumn{3}{c}{MyoPS(5 scribble)}& \multicolumn{4}{c}{MSCMRseg(5 scribble)}\\
    \cmidrule(lr){4-5}\cmidrule(lr){7-9}\cmidrule(lr){10-13}
   & & &  sup & \multicolumn{1}{c}{reg}&& Scar & Edema & \multicolumn{1}{c|}{Avg} & LV & MYO & RV &\multicolumn{1}{c}{Avg}\\
				\hline
		      \multicolumn{1}{c|}{\#1}&$\checkmark$&$\times$&$\times$&$\times$&\multicolumn{1}{|c|}{$\times$}&.310$\pm$.257&.153$\pm$.090&\multicolumn{1}{c|}{.231$\pm$.199}&.373$\pm$.346&.337$\pm$.254&.139$\pm$.143&\multicolumn{1}{c}{.283$\pm$.264}\\
    	\multicolumn{1}{c|}{\#2}&$\checkmark$&$\checkmark$&$\times$&$\times$&\multicolumn{1}{|c|}{$\times$}&.348$\pm$.189&.531$\pm$.106&\multicolumn{1}{c|}{.440$\pm$.177}&.700$\pm$.234&.416$\pm$.190&.501$\pm$.460&\multicolumn{1}{c}{.539$\pm$.319}\\	
	\multicolumn{1}{c|}{\#3}&$\checkmark$&$\checkmark$&$\checkmark$& $\times$&\multicolumn{1}{|c|}{$\times$}&\underline{.480$\pm$.074}&\underline{.625$\pm$.197}&\multicolumn{1}{c|}{\underline{.552$\pm$.160}}&\underline{.823$\pm$.104}&\underline{.717$\pm$.117}&\underline{.508$\pm$.419}&\multicolumn{1}{c}{\underline{.683$\pm$.275}}\\	
	\multicolumn{1}{c|}{\textbf{\#4}}&$\checkmark$&$\checkmark$&$\checkmark$&$\checkmark$&\multicolumn{1}{|c|}{$\times$}&\bf{.518$\pm$.081}&\bf{.630$\pm$.128}&\multicolumn{1}{c|}{\bf{.574$\pm$.117}}&\textbf{.922$\pm$.050}&\textbf{.799$\pm$.095}&\textbf{.696$\pm$.267}&\multicolumn{1}{c}{\textbf{.805$\pm$.181}}\\
     \cdashline{1-13}
 	\multicolumn{1}{c|}{\#5}&$\checkmark$&$\checkmark$&-&-&\multicolumn{1}{|c|}{$\checkmark$}&.242$\pm$.166&.386$\pm$.150&\multicolumn{1}{c|}{.314$\pm$.168}&.581$\pm$.094&.462$\pm$.101&.346$\pm$.265&\multicolumn{1}{c}{.463$\pm$.188}\\
	\hline
   &&&&&\multicolumn{1}{c|}{}&\multicolumn{3}{c|}{MyoPS(5 scribble)}& \multicolumn{4}{c}{ACDC(5 scribble)}\\
   \hline
\multicolumn{1}{c|}{\#1}&$\checkmark$&$\times$&$\times$&$\times$&\multicolumn{1}{|c|}{$\times$}&.310$\pm$.257&.153$\pm$.090&\multicolumn{1}{c|}{.231$\pm$.199}&.585$\pm$.249&.466$\pm$.147&.279$\pm$.176&\multicolumn{1}{c}{.443$\pm$.231}\\
    	\multicolumn{1}{c|}{\#2}&$\checkmark$&$\checkmark$&$\times$&$\times$&\multicolumn{1}{|c|}{$\times$}&.348$\pm$.189&.531$\pm$.106&\multicolumn{1}{c|}{.440$\pm$.177}&.580$\pm$.240&.544$\pm$.183&\underline{.486$\pm$.294}&.537$\pm$.244\\	
	\multicolumn{1}{c|}{\#3}&$\checkmark$&$\checkmark$&$\checkmark$& $\times$&\multicolumn{1}{|c|}{$\times$}&\underline{.449$\pm$.059}&\underline{.644$\pm$.113}&\multicolumn{1}{c|}{\underline{.546$\pm$.134}}&\underline{.803$\pm$.161}&\underline{.686$\pm$.124}&.462$\pm$.303&\multicolumn{1}{c}{\underline{.650$\pm$.253}}\\	
	\multicolumn{1}{c|}{\textbf{\#4}}&$\checkmark$&$\checkmark$&$\checkmark$&$\checkmark$&\multicolumn{1}{|c|}{$\times$}&\textbf{.543$\pm$.170}&\textbf{.588$\pm$.194}&\multicolumn{1}{c|}{\textbf{.566$\pm$.173}}&\textbf{.829$\pm$.145}&\textbf{.743$\pm$.127}&\textbf{.728$\pm$.131}&\multicolumn{1}{c}{\textbf{.767$\pm$.140}}\\
     \cdashline{1-13}
 	\multicolumn{1}{c|}{\#5}&$\checkmark$&$\checkmark$&-&-&\multicolumn{1}{|c|}{$\checkmark$}&.179$\pm$.173&.258$\pm$.227&\multicolumn{1}{c|}{.219$\pm$.195}&.783$\pm$.142&.473$\pm$.129&.143$\pm$.133&\multicolumn{1}{c}{.466$\pm$.295}\\
	\hline
   \hline
	\end{tabular}}
	\end{center}
 \label{tab:component_ablations}
\end{table}	
\subsection{Ablation study}
We verified the effectiveness of ModelMix components on two task combinations, \emph{i.e}, MyoPS \& MSCMRseg and MyoPS \& ACDC. All models are trained using five scribbles and then evaluated on the validation set. \emph{Five} ablated models are implemented, comprising the partial cross-entropy loss ($\mathcal{L}_{\text{pce}}$) computed from annotated pixels, Mix invariant loss ($\mathcal{L}_{\text{inv}}$), and the vicinal loss ($\mathcal{L}_{\text{vicinal-sup}},\mathcal{L}_{\text{vicinal-reg}}$). We also compared the results to the baseline trained with shared encoder (Share Enc). Details are summarized in Table \ref{tab:component_ablations}.
\\
\\
\noindent\textbf{Components of ModelMix:} When integrating Mix-based augmentations, there is a significant improvement in the performance of model \#2 compared to model \#1, with an average Dice increase of 20.9\%, 25.6\%, and 9.4\%  on MyoPS, MSCMRseg, and ACDC, respectively.
When combined with Supervised vicinal loss ($\mathcal{L}_{\text{vicinal-sup}}$), Model \#3 obtained remarkable performance gain on all datasets, demonstrating that learning from complementary tasks can promote the performance of each individual task.
When leveraging unlabeled pixels with unsupervised vicinal loss ($\mathcal{L}_{\text{vicinal-reg}}$) , the average Dice Scores of model \#4 on MyoPS, MSCMR, and ACDC are further boosted to 57.4\%, 80.5\%, and 76.7\% , respectively.
\\
\\
\noindent\textbf{Combination of Tasks:}
We combine MyoPS with MSCMRseg and ACDC respectively and compared their segmentation performance.
Note that images from MSCMRseg and MyoPS both contain enhanced pathological information, while ACDC only includes structural information.
Therefore, the complementarity between MyoPS and MyoPS is greater than the combination of MyoPS and ACDC.
As expected, the performance improvement of MyoPS with MSCMR is slightly better than that of MyoPS with ACDC, improving by 13.4\% and 12.6\% over separate trained models, respectively.
\\
\\
\noindent\textbf{Comparison with shared encoder:}
In \zkreftb{tab:component_ablations}, we further compare our method with the shared encoder baseline of Model \#5. 
Employing a shared encoder (Share Enc) for tasks, Model \#5 exhibits inferior performance compared to Model \#2, which is trained using separate encoders.
This disparity arises because when training supervision is limited, employing a single encoder to generalize across various tasks poses greater challenges to model training. Leveraging separate encoders and constructed virtual models in the vicinity between individual models, our approach (Model \#4) notably surpasses both Model \#2 and Model \#5 by large margins.

\begin{table}[!t]
	\caption{The 5-scribble supervised segmentation results on MSCMRseg dataset.}\label{tab:tab_mscmr}
	
	\begin{center}
		\resizebox{\textwidth}{!}{
		\begin{tabular}{cccccccccccccccccccc}
                \hline
			\hline
	\multirow{2}{*}{Methods}& \multicolumn{4}{c}{Dice} & \multicolumn{4}{c}{HD(mm)}\\
 \cmidrule(lr){2-5}\cmidrule(lr){6-9}
    & LV & MYO & RV &\multicolumn{1}{c|}{Avg} &LV & MYO & RV & \multicolumn{1}{c}{Avg} \\
				\hline
 	\multicolumn{1}{l|}{PCE}&.454$\pm$.271&.356$\pm$.182&.102$\pm$.072&\multicolumn{1}{c|}{.304$\pm$.241}&\underline{111.87$\pm$14.40}&\underline{109.72$\pm$20.30}&123.62$\pm$19.33&115.07$\pm$16.02\\
    	\multicolumn{1}{l|}{Mixup}&.440$\pm$.102&.310$\pm$.127&.021$\pm$.013&\multicolumn{1}{c|}{.257$\pm$.200}&259.42$\pm$14.18&210.00$\pm$12.37&251.98$\pm$15.6&240.47$\pm$25.96\\	
	\multicolumn{1}{l|}{Cutout}&.315$\pm$.103&.307$\pm$.153&.166$\pm$.110&\multicolumn{1}{c|}{.263$\pm$.139}&259.42$\pm$14.18&240.06$\pm$16.38&252.18$\pm$15.42&250.56$\pm$17.04\\	
        \multicolumn{1}{l|}{CycleMix}&.517$\pm$.086&.421$\pm$.108&.007$\pm$.007&\multicolumn{1}{c|}{.315$\pm$.237}&213.20$\pm$35.65&151.36$\pm$55.12&260.56$\pm$12.66&208.37$\pm$58.88\\	
        \multicolumn{1}{l|}{ShapePU}&.758$\pm$.191&.567$\pm$.168&.059$\pm$.026&\multicolumn{1}{c|}{.461$\pm$.331}&209.04$\pm$16.09&234.08$\pm$18.15&237.86$\pm$14.13&226.99$\pm$20.45\\	
        \multicolumn{1}{l|}{WSL4}&\underline{.809$\pm$.079}&\underline{.653$\pm$.109}&\underline{.599$\pm$.261}&\multicolumn{1}{c|}{\underline{.687$\pm$.191}}&140.95$\pm$69.06&147.74$\pm$59.93&\textbf{95.07$\pm$60.53}&\underline{127.92$\pm$67.49}\\	
        \multicolumn{1}{l|}{\textbf{w/ MyoPS}}&\textbf{.875$\pm$.077}&\textbf{.754$\pm$.079}&\textbf{.722$\pm$.201}&\multicolumn{1}{c|}{\bf{.784$\pm$.145}}&\bf{78.05$\pm$16.11}&\bf{69.85$\pm$30.45}&\underline{99.20$\pm$46.81}&\bf{82.36$\pm$35.09}\\	
	\hline
        \multicolumn{1}{l|}{FullSup-UNet}&.775$\pm$.158&.604$\pm$.147&.572$\pm$.207&\multicolumn{1}{c|}{.651$\pm$.191}&23.50$\pm$21.79&34.03$\pm$19.25&81.29$\pm$11.29&46.27$\pm$30.91\\	
        \multicolumn{1}{l|}{FullSup-nnUNet}&.885$\pm$.085&.757$\pm$.147&.757$\pm$.201&\multicolumn{1}{c|}{.799$\pm$.160}&21.48$\pm$29.68&13.50$\pm$12.99&18.27$\pm$12.51&17.75$\pm$19.87\\	
        \hline
        \hline
		\end{tabular}}
	\end{center}
\end{table}	
\begin{figure}[t]
    \centering
    \includegraphics[width=\textwidth]{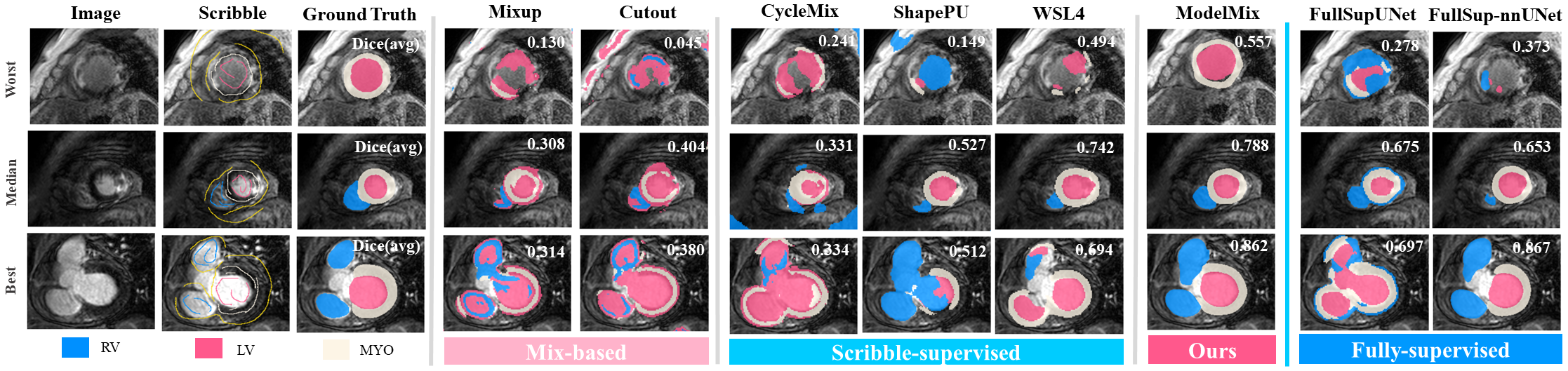}\\[-1ex] 
    \caption{The qualitative comparison on MSCMRseg dataset. The three images are the worst, median and best cases selected by the average Dice Score.}
    \label{fig:samples}
\end{figure}

\subsection{Performance and Comparisons}
We compare our method against three sets of \emph{10} baselines. 
The first set includes six scribble-guided models, \emph{i.e.}, partial cross-entropy (PCE)~\cite{tang2018normalized}, Mixup~\cite{zhang2018mixup}, Cutout~\cite{devries2017cutout}, CycleMix~\cite{zhang2022cyclemix}, ShapePU~\cite{zhang2022shapepu}, and WSL4~\cite{luo2022scribble}.
The second set comprises two semi-supervised methods of positive and unlabeled learning, including CVIR~\cite{garg2021mixture} and nnPU~\cite{NIPS2017_7cce53cf}. These methods are provided with the additional information of label class ratio, and then adapted for scribble supervised segmentation.
Finally, we include results from fully supervised UNet~\cite{baumgartner2017exploration} (FullSup-UNet) and fully supervised nnUNet~\cite{isensee2021nnu} (FullSup-nnUNet) for reference.
\\
\\
\noindent\textbf{Comparison with scribble-supervised methods:} Table \ref{tab:tab_myops} presents the quantitative comparisons on the MSCMR dataset.
When mixing with the MyoPS model, our ModelMix(w/ MyoPS) significantly outperforms all other scribble-supervised methods by an average of 9.7\% in Dice Score.
This is affirmed by the qualitative results of \zkreffig{fig:samples}, which visualizes the results of the worst, median, and best cases by the average Dice Scores of all compared methods.
\\
\begin{table*}[!t]
	\caption{The 5-scribble guided segmentation results on MyoPS dataset.}\label{tab:tab_myops}
	\centering
		\resizebox{\linewidth}{!}{
			\begin{tabular}{ccccccccccc}
				\hline
                    \hline
				\multirow{2}{*}{Methods}&\multirow{2}{*}{Ratio}&\multirow{2}{*}{ACDC}&\multirow{2}{*}{MSCMR}&\multicolumn{3}{c}{Dice}& \multicolumn{3}{c}{HD(mm)}\\
				\cmidrule(lr){5-7}\cmidrule(lr){8-10}
				&&&&\multicolumn{1}{c}{Scar} & Edema & \multicolumn{1}{c|}{Avg} &Scar & Edema &\multicolumn{1}{c}{Avg}\\
				\hline
				\multicolumn{1}{c}{PCE}&\multicolumn{1}{|c|}{$\times$}&$\times$&\multicolumn{1}{c|}{$\times$}&.242$\pm$.170&.122$\pm$.077&\multicolumn{1}{c|}{.182$\pm$.144}&76.22$\pm$37.24&124.89$\pm$21.27&100.55$\pm$38.77\\ 
                \multicolumn{1}{c}{CVIR~\cite{garg2021mixture}}&\multicolumn{1}{|c|}{$\checkmark$}&$\times$&\multicolumn{1}{c|}{$\times$}&.288$\pm$.191&.085$\pm$.034&\multicolumn{1}{c|}{.186$\pm$.170}&45.01$\pm$18.44&125.27$\pm$20.83&85.14$\pm$45.04\\	
                \multicolumn{1}{c}{nnPU~\cite{NIPS2017_7cce53cf}}&\multicolumn{1}{|c|}{$\checkmark$}&$\times$&\multicolumn{1}{c|}{$\times$}&.290$\pm$.166&.236$\pm$.078&\multicolumn{1}{c|}{.263$\pm$.131}&126.51$\pm$35.27&125.05$\pm$20.69&125.78$\pm$28.55\\
                \cdashline{1-10}
			\multirow{3}{*}{\textbf{ours}}&\multicolumn{1}{|c|}{$\times$}&$\checkmark$&\multicolumn{1}{c|}{$\times$}&.455$\pm$.251&.518$\pm$.140&\multicolumn{1}{c|}{.487$\pm$.203}&77.52$\pm$34.68&81.85$\pm$28.37&\multicolumn{1}{c}{79.69$\pm$31.35}\\
			    &\multicolumn{1}{|c|}{$\times$}&$\times$&\multicolumn{1}{c|}{$\checkmark$}&\textbf{.488$\pm$.263}&\textbf{.575$\pm$.147}&\multicolumn{1}{c|}{\textbf{.532$\pm$.215}}&\underline{70.18$\pm$34.19}&\underline{75.57$\pm$28.13}&\multicolumn{1}{c}{\underline{72.87$\pm$31.02}}\\
    			&\multicolumn{1}{|c|}{$\times$}&$\checkmark$&\multicolumn{1}{c|}{$\checkmark$}&\underline{.474$\pm$.269}&\underline{.545$\pm$.158}&\multicolumn{1}{c|}{\underline{.509$\pm$.221}}&\textbf{41.26$\pm$20.38}&\textbf{46.63$\pm$18.71}&\multicolumn{1}{c}{\textbf{43.95$\pm$19.50}}\\
               \hline
			    \multicolumn{1}{c}{FullSup-UNet}&\multicolumn{1}{|c|}{-}&$\times$&\multicolumn{1}{c|}{$\times$}&.423$\pm$.253&.445$\pm$.149&\multicolumn{1}{c|}{.434$\pm$.205}&117.61$\pm$35.08&119.13$\pm$22.7&118.37$\pm$29.17\\ 
			    \multicolumn{1}{c}{FullSup-nnUNet}&\multicolumn{1}{|c|}{-}&$\times$&\multicolumn{1}{c|}{$\times$}&.496$\pm$.252&.563$\pm$.141&\multicolumn{1}{c|}{.529$\pm$.204}&43.86$\pm$37.27&45.14$\pm$33.86&\multicolumn{1}{c}{44.50$\pm$35.15}\\
				\hline
                    \hline
		\end{tabular}}
\end{table*}
\begin{figure}[t]
    \centering
    \includegraphics[width=\textwidth]{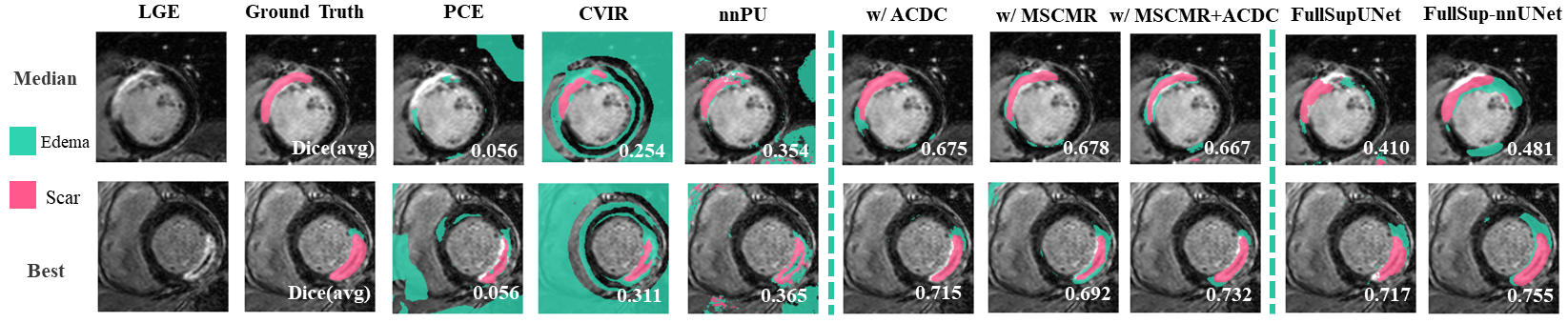}\\[-1ex] 
    \caption{The qualitative comparison on MyoPS dataset. The two images are the median and best cases selected by the average Dice Score.}
    \label{fig:myops_samples}
\end{figure}

\noindent\textbf{Comparison with semi-supervised methods:} 
Table~\ref{tab:tab_myops} summarizes the results on the MyoPS dataset. Considering that traditional pseudo-label based methods (\emph{i.e.}, WSL4) are used to segment regular structures and fail to converge on this challenging task, we compare our methods against the adapted semi-supervised methodologies such as CVIR~\cite{garg2021mixture} and nnPU~\cite{NIPS2017_7cce53cf}, with additional input of category mix ratios. Mixing with model parameters of ACDC and MSCMRseg separately, our ModelMix obtains remarkable performance gain, exceeding the second-best comparison methods by 22.4\% and 26.9\% in Dice, respectively.  
When combined with both ACDC and MSCMRseg, ModelMix learns robust anatomical priors and demonstrates the competitive performance on both Dice and HD. \zkreffig{fig:myops_samples} visualizes the median and best cases selected by average Dice of all compared methods.
One can observe that our ModelMix achieves more realistic segmentation results than other methods when mixed with complementary tasks (w/ ACDC, MSCMR, ACDC+MSCMR).

\section{Conclusion}
We introduce ModelMix, a simple and model-agnostic approach to blend model parameters of complementary tasks. Our method minimize the vicinal risk of virtual models, which are constructed through linear interpolation of encoder convolutional parameters and their corresponding features. Extensive evaluations on three public datasets demonstrate that ModelMix achieves state-of-the-art performance in the challenging task of few-shot scribble-supervised segmentation.
%
%
\bibliographystyle{splncs04}
\bibliography{ref}
\end{document}


\title{Supplementary Material of ModelMix}
%
%
%
%
\maketitle              

\subsection{Supervision sensitivity}
We verified the supervised sensitivity of the proposed ModelMix by training the model using 20\% scribble annotated images and 80\% unlabeled images, as well as 100\% scribble labeled images. \zkreftb{sensitivity_1} and \zkreftb{sensitivity_2} summarizes the experiment results on MyoPS and MSCMRseg dataset. 
\\
\\
\textbf{Irregular pathology segmentation of MyoPS:}
One can observe that our ModelMix consistently outperforms semi-supervised benchmarks with both 20\% and 100\% scribble annotations on MyoPS datasets. 
As shown in \zkreftb{sensitivity_1}, by mixing with the model parameters of MSCMR dataset, our ModelMix obtains remarkable performance gain of 26.9\% and 25.0\% Dice with 20\% and 100\% scribble annotations, respectively. 
These results demonstrate the robustness of the proposed ModelMix in the situation of different supervision amount.

\begin{table*}[htb]
\renewcommand{\thetable}{\uppercase\expandafter{\romannumeral1}}
	\caption{Supervision sensitivity: Irregular pathology segmentation on MyoPS dataset.}\label{sensitivity_1}
	\centering
		\resizebox{0.8\linewidth}{!}{
			\begin{tabular}{ccccccccc}
				\hline
				\multirow{2}{*}{Methods}&\multirow{2}{*}{Ratio}&\multicolumn{3}{c}{20\% scribbles} &\multicolumn{3}{c}{100\% scribbles}\\
    \cmidrule{3-5} \cmidrule{6-8}
    &&Scar& Edema & \multicolumn{1}{c}{Avg}&Scar& Edema & \multicolumn{1}{c}{Avg}\\
				\hline
				\multicolumn{1}{l|}{PCE}&\multicolumn{1}{c|}{$\times$}&.242$\pm$.170&.122$\pm$.077&\multicolumn{1}{c|}{.182$\pm$.144}&.504$\pm$.213&.057$\pm$.022&.281$\pm$.271\\ 
                \multicolumn{1}{l|}{CVIR}&\multicolumn{1}{c|}{$\checkmark$}&.288$\pm$.191&.085$\pm$.034&\multicolumn{1}{c|}{.186$\pm$.170}&.505$\pm$.214&.080$\pm$.031&.293$\pm$.263\\		
                \multicolumn{1}{l|}{nnPU}&\multicolumn{1}{c|}{$\checkmark$}&\underline{.290$\pm$.166}&\underline{.236$\pm$.078}&\multicolumn{1}{c|}{\underline{.263$\pm$.131}}&\underline{.530$\pm$.241}&\underline{.085$\pm$.035}&\underline{.308$\pm$.282}\\
			    \multicolumn{1}{l|}{\textbf{w/ MSCMR}}&\multicolumn{1}{c|}{$\times$}&\textbf{.488$\pm$.263}&\textbf{.575$\pm$.147}&\multicolumn{1}{c|}{\textbf{.532$\pm$.215}}&\textbf{.541$\pm$.268}&\textbf{.575$\pm$.214}&\textbf{.558$\pm$.240}\\
                \cdashline{1-8}
			    \multicolumn{1}{l|}{FullSup-UNet}&\multicolumn{1}{c|}{-}&.423$\pm$.253&.445$\pm$.149&\multicolumn{1}{c|}{.434$\pm$.205}&.537$\pm$.232&.659$\pm$.135&.633$\pm$.202\\  
			    \multicolumn{1}{l|}{FullSup-nnUNet}&\multicolumn{1}{c|}{-}&.496$\pm$.252&.563$\pm$.141&\multicolumn{1}{c|}{.529$\pm$.204}&.610$\pm$.169&.651$\pm$.246&.630$\pm$.209\\
				\hline
                \hline
		\end{tabular}}
\end{table*}
\\

\noindent\textbf{Regular ventrical segmentation of MSCMRseg:} By changing the supervision amount from 20\% to 100\%, we also verify the supervision sensitivity of proposed ModelMix on MSCMRseg dataset.
As illustrated in \zkreftb{sensitivity_2}, the proposed ModelMix succeeds in both scenarios, with evident improvement in average Dice by 9.7\% and 2.7\%, respectively. 

\begin{table}
\renewcommand{\thetable}{\uppercase\expandafter{\romannumeral2}}
	\caption{Supervision sensitivity: regular ventrical segmentation on MSCMRseg dataset.}\label{sensitivity_2}
	\centering
		\resizebox{0.8\textwidth}{!}{
		\begin{tabular}{cccccccccccccccccccc}
                \hline
			\hline
	\multirow{2}{*}{Methods}& \multicolumn{4}{c}{20\% scribbles} & \multicolumn{4}{c}{100\% scribbles}\\
 \cmidrule(lr){2-5}\cmidrule(lr){6-9}
    & LV & MYO & RV &\multicolumn{1}{c|}{Avg} &LV & MYO & RV & \multicolumn{1}{c}{Avg} \\
				\hline
    	\multicolumn{1}{l|}{Mixup}&.440$\pm$.102&.310$\pm$.127&.021$\pm$.013&\multicolumn{1}{c|}{.257$\pm$.200}&.483$\pm$.09&.466$\pm$.080&.455$\pm$.134&.468$\pm$.102\\	
	\multicolumn{1}{l|}{Cutout}&.315$\pm$.103&.307$\pm$.153&.166$\pm$.110&\multicolumn{1}{c|}{.263$\pm$.139}&468$\pm$.076&.642$\pm$.132&.694$\pm$.146&.602$\pm$.154\\
        \multicolumn{1}{l|}{CycleMix}&.517$\pm$.086&.421$\pm$.108&.007$\pm$.007&\multicolumn{1}{c|}{.315$\pm$.237}&.872$\pm$.060&.734$\pm$.048&.787$\pm$.073&.798$\pm$.083\\
        \multicolumn{1}{l|}{ShapePU}&.758$\pm$.191&.567$\pm$.168&.059$\pm$.026&\multicolumn{1}{c|}{.461$\pm$.331}&.880$\pm$.046&.785$\pm$.080&\underline{.833$\pm$.087}&.833$\pm$.082\\
        \multicolumn{1}{l|}{WSL4}&\underline{.809$\pm$.079}&\underline{.653$\pm$.109}&\underline{.599$\pm$.261}&\multicolumn{1}{c|}{\underline{.687$\pm$.191}}&\underline{.902$\pm$.040}&\underline{.815$\pm$.033}&.828$\pm$.101&\underline{.848$\pm$.076}\\
        \multicolumn{1}{l|}{\textbf{w/ MyoPS}}&\textbf{.875$\pm$.077}&\textbf{.754$\pm$.079}&\textbf{.722$\pm$.201}&\multicolumn{1}{c|}{\bf{.784$\pm$.145}}&\textbf{.919$\pm$.036}&\textbf{.842$\pm$.035}&\textbf{.865$\pm$.063}&\textbf{.875$\pm$.056}\\
	\cdashline{1-9}
        \multicolumn{1}{l|}{FullSup-UNet}&.775$\pm$.158&.604$\pm$.147&.572$\pm$.207&\multicolumn{1}{c|}{.651$\pm$.191}&.917$\pm$.046&.813$\pm$.058&.750$\pm$.162&.827$\pm$.122\\
        \multicolumn{1}{l|}{FullSup-nnUNet}&.885$\pm$.085&.757$\pm$.147&.757$\pm$.201&\multicolumn{1}{c|}{.799$\pm$.160}&.909$\pm$.049&.880$\pm$.027&.902$\pm$.047&.907$\pm$.044\\
        \hline
        \hline
		\end{tabular}}
\end{table}	
\subsection{The details of compared methods}
We compare our ModelMix to ten methods, the detailed information are summarized as follows:
\begin{itemize}
\item[1)] \textcolor{black}{\textit{PCE:} UNet backbone trained with the PCE loss ($\mathcal{L}_{\text{pce}}$)}.

\item[2)] \textcolor{black}{\textit{Mixup:} UNet backbone trained with mixup augmentation and PCE loss.}

\item[3)] \textcolor{black}{\textit{Cutout:} UNet backbone trained with cutout augmentation and PCE loss.}

\item[4)] \textcolor{black}{\textit{WSL4}: We adopted code of WSL4 released by the authors via \href{https://github.com/HiLab-git/PyMIC/blob/master/pymic/net_run/weak_sup/wsl_dmpls.py}{\textcolor{black}{https://github.com/HiLab-git/PyMIC/blob/master/pymic}}}.

\item[5)] \textcolor{black}{\textit{CycleMix}: We use the code released via \href{https://github.com/BWGZK/CycleMix}{\textcolor{black}{https://github.com/BWGZK/CycleMix}}. }

\item[6)] \textcolor{black}{\textit{ShapePU}: We leverage the code released via \href{https://github.com/BWGZK/ShapePU}{\textcolor{black}{https://github.com/BWGZK/ShapePU}}. }

\item[7)] \textcolor{black}{\textit{CVIR}: We use the released code from \href{https://github.com/acmi-lab/PU_learning}{\textcolor{black}{https://github.com/acmi-lab/PU\_learning}} to implement the technique of Conditional Value Ignoring Risk (CVIR). Since CVIR is developed on the condition that mixture ratio is known, we provide it with ground truth ratio for model training.
Given that CVIR is proposed for classification tasks, we apply it to each individual pixel classification task to achieve the pixel-level segmentation.}
\item[8)] \textcolor{black}{\textit{nnPU}: The reproduced code via \href{https://github.com/kiryor/nnPUlearning}{\textcolor{black}{https://github.com/kiryor/nnPUlearning}} is adopted for our implementation of PU loss. Since nnPU is designed for classification tasks, we adapt it to segmentation tasks by applying it to each pixel.}
\item[9)] \textcolor{black}{\textit{FullSup-UNet}: The fully supervised UNet trained with the cross-entropy loss calculated with full annotations.}
\item[10)] \textcolor{black}{\textit{FullSup-nnUNet}: We adopot the code released via \href{https://github.com/MIC-DKFZ/nnUNet}
{\textcolor{black}{https://github.com/MIC-DKFZ/nnUNet}}.}
\end{itemize}

\begin{figure}[!t]
    \centering
    \renewcommand{\thefigure}{\uppercase\expandafter{\romannumeral2}}
    \includegraphics[width=0.7\textwidth]{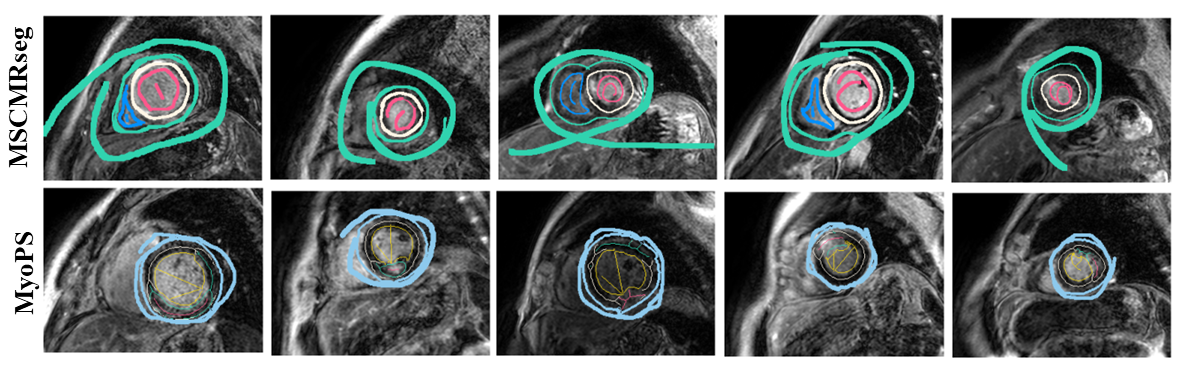}\\[-1ex] 
    \caption{The visualization of scribbles from MSCMRseg and MyoPS dataset.}
    \label{fig:scribble}
\end{figure}
\subsection{Visualization of scribbles:} The typical scribble annotations of MSCMRseg and MyoPS dataset are visualized in \zkreffig{fig:scribble}.

%